# Legal entity recognition in an agglutinating language and document connection network for EU Legislation and EU/Hungarian Case Law

György Görög
*Dept. of Knowledge Management*
*Ministry of Defence*
*Electronics, Logistics & Property Management Co.*
Budapest, Hungary
gorog.gyorgy@hmei.hu
ORCID: 0000-0003-3007-2622

Péter Weisz
*Dept. of Knowledge Management*
*Ministry of Defence*
*Electronics, Logistics & Property Management Co.*
Budapest, Hungary
weisz.peter@hmei.hu
ORCID: 0000-0002-2011-2697

*Abstract*—We have developed an application aiming at federated search for EU and Hungarian legislation and jurisdiction. It now contains above 1 million documents, with daily updates. The database holds documents downloaded from the EU sources EUR-Lex and Curia Online as well as public jurisdiction documents from the Constitutional Court of Hungary and The National Office for The Judiciary. The application is termed Justeus. Justeus provides comprehensible search possibilities. Besides free text and metadata (dropdown list) searches, it features hierarchical data structures (concept hierarchy trees) of directory codes and classification as well as subject terms. Justeus collects all links of a particular document to other documents (court judgements citing other case law documents as well as legislation, national court decisions referring to EU regulation etc.) as tables and directed graph networks. Choosing a document, its relations to other documents are visualized in real time as a network. Network graphs help in identifying key documents influencing or referred by many other documents (legislative and/or jurisdictive) and sets of documents predominantly referring to each other (citation networks).

## I. Introduction

European Union (EU) public legislation and case-law documents are freely available through at least two search engines: EUR-Lex and Curia Online. Besides commercial applications, Hungarian (HU) case-law documents are published at the National Office for The Judiciary (OBH) and the Constitutional Court (AB) home pages. However, these sources are searchable only separately, and cross-references (e.g. HU documents referring to an EU directive or EU Curia decision, or even a Hungarian court ruling being challenged before AB) are hidden in the texts. In addition, some of these sources are poorly equipped with metadata and may suffer from typos, scanner/OCR anomalies and severe entity loss due to overenthusiastic anonymization.

Our goal was to provide professional and lay users with a single source of the above mentioned document types, later to be completed with Hungarian legislation. With EU rules having primacy over national regulation in some areas, judges, lawyers and laypersons need an interconnected, single step search engine focusing on legal references between documents, especially HU–EU relations.

A federated, multi-national, pay-for EU case-law search engine (EuroCases) is published by Apis Europe. Our application puts more emphasis on EU legislation, and elaborates only on HU and EU text corpora. Justeus has been published and is currently in beta testing phase. At the time of this writing, it is available free of charge against registration requiring a valid e-mail address only; no personal details are asked for. User activity is logged for analytical purposes.

## II. Workflow Overview

### A. Downloaders

Each public data source has its own downloader. EUR-Lex notifies about changed and new documents through its SOAP web service, although it sometimes misses actual document modifications. Obtaining Celex (the EUR-Lex document identifier) and Cellar identifiers from this web service, the downloader gets the XML metadata from Cellar, the common content and metadata repository of the EU Publications Office. HTML (text) content is downloaded from Cellar, EUR-Lex, the Publications Office, or the Official Journal, wherever available (varies). EU contents are downloaded in Hungarian, English, German and French.

EU Curia Online provides case and document listings and individual rulings are accessed through them.

AB provides a single-page, concise HTML representation of the decision text and all metadata they publish.

OBH has a difficult to penetrate download interface that leads to a limited set of metadata and RTF contents.

Downloaded metadata and contents are stored in an MS SQL Server database. In all cases, if the extractors find a

The Justeus research and development project was financed with the support of the Hungarian National Fund for Research, Development and Innovation (Nemzeti Kutatási, Fejlesztési és Innovációs Alap) with code PIAC-13-1-2013-0117.

reference to a document missing from the SQL tables, the downloader tries to find it in the original source. This results in an approx. 10% higher completeness, depending on the source. Currently the repository contains over 1M EU and approx. 150,000 HU documents.

### B. Identifiers

The main identifier is Celex, originally defined for EUR-Lex documents. We generate Celex numbers for EU Curia documents lacking one; this facilitates pairing of EUR-Lex and Curia Online versions. For any other document lacking a Celex number, most notably Hungarian ones, a Celex-like identifier is also created, following EUR-Lex generation rules as far as reasonable. AB maintains a GUID identifier for cases; this remains unchanged when a decision has been reached. The OBH ruling identifier is unique only with regard of the given court (each court having an integer id).

The European Case Law Identifier (ECLI) and case/decision/legislation numbers are also used to find references in texts.

### C. Extractors

Extraction of metadata from structured forms like XML and text starts with finding and formatting metadata provided by the data sources. Raw data are normalized, being aided by authority tables (e.g. EU institutions' names and codes) where available. There are also cleaning steps to avoid typos, OCR errors, inconsequent XML and HTML DOM structures (e.g. unescaped < and > signs in document titles) and similar anomalies. Then, texts are submitted to entity recognition, entities being references to other documents, judges, applicant and defendant, their representatives, subjects of the case and the like.

Over 130 metadata types are maintained. Metadata values are provided to the user as table rows; references to other documents in texts are transformed to links. Acronyms are presented as html <abbrev> tags with their full forms as titles.

### D. Indexing

The search engine is a Micro Focus IDOL with a custom-developed Hungarian stemmer.

At present, only Hungarian language document contents are submitted for stemmed text indexing. Search modes include exact phrase, all words, any word, proximity, and freely editable expressions (expert search). Search uses synonyms if desired; synonyms come from Eurovoc, the EU's multilingual thesaurus [1]. Search terms are highlighted in found document text.

Metadata (document descriptors) are submitted for fast search indexing ("match fields" or "index fields" in IDOL terminology) or as nonindexed, nonsearchable fields to be displayed only.

### E. User interface

The user interface is custom-built. The search interface is an Asp.net application with a high number of dropdown menus to facilitate metadata search besides free-text search. The document display front-end is an Angular application served by a .NET Core back-end. The network diagram uses vis.js [2] for layout and visualization.

### F. Production setup

The production version of Justeus resides on an Azure account with IDOL installed on a virtual machine, the user interface served by an IIS service and the SQL tables maintained on an Azure SQL service. Justeus is available at www.justeus.eu.

## III. THE HUNGARIAN LANGUAGE

Hungarian is atypical within Uralic languages and has no close relatives. It is an agglutinating language (inflecting mainly by suffixes) with a high morphological variation [3], an excessive usage of accented characters and two/three character consonant phonemes as well as a variable word order (but Saxon genitive only.) In turn, there is no grammatical gender.

Efficient free-text search requires a stemmer, and a good stemmer is even more essential for an agglutinating language. We found existing stemmers unsatisfactory for our purposes and teamed up with a language technology group that developed their stemmer further to our needs. Development is continuous in this field.

Entity recognition is hampered by the extreme morphological diversity if a stemmer is not used. Suffixes often modify the stem itself by leaving out or inserting a letter or changing the accent. Sensitive entity recognition is best performed on accent-free text, but then the accents should be restored for presentation to the user. However, vowel harmony requirements make it difficult to predict the exact suffix for a stem. We have not been able to achieve high recall, high precision entity extraction without normalized entity (authority) lists, at least when the extracted terms are presented as e.g. table rows or dropdown items and should be in familiar form. A few erroneous entries may ruin the credibility and aesthetics of a dropdown list.

In legal texts, variability is somewhat smaller then in general texts; word order is especially more predictable, with e.g. a court name always preceding the case number. Legal language extraction, at least in our hands, usually does not deal with verbs. Compounds usage is inconsequent; two-word constructs may be joined with a hyphen, concatenated or left separated by a space; so oftentimes, we attempt to find entities in concatenated sentences (spaces and hyphens removed).

## IV. ENTITY RECOGNITION

Although EU data sources provide extensive metadata, we aim at finding at least a subset of them (document references) in the texts to facilitate reading and access to these referenced documents. HU data sources provide less detailed metadata so we have to rely more on text analysis to find relevant information.

These extraction processes are usually semi-automatic [4] or target a limited set of entities [5]. Apis Europe offers EULinksChecker Add-in Tools to recognize EU references (and Eurovoc terms) in text in some languages.

We implemented a fully automatic approach that of course still requires human supervision.

### A. Recognition of document references

Legal documents are usually referred to in legal texts in a strict format; for example, EU case law (case numbers)

have a characteristic form (e.g. C-18/16) that is not easy to mix up with other references. Still, there are variations and disturbing similarities among numbers from different sources; for instance, both EU regulations and AB decisions use a 12/2016 type format (and EU directives use a 2016/12 format for distinction).

Therefore, in most of the cases, extraction relies on the document reference itself and the text environment alike. For instance, HU court decisions have an unmistakable format like "4.K.27.207/2015/12." They are, however, not unique as two courts may assign the same number to their respective decisions. Therefore, in text, we have to observe the court's name as well (in Hungarian it always precedes the decision number) and look it up in the database together.

A typical sentence often contains more than one court names ("The Supreme Court approves the Budapest Court's 4.K.27.207/2015/12 judgement") or two or more decision numbers in a single sentence, belonging to one or two courts. Similarly, two or more articles of an EU treaty may be referenced in a list ("Article 7, Articles 13 to 19, Article 48(2) to (5), and Articles 49 and 50 of the Treaty on European Union shall apply to this Treaty"). The extractors will store a reasonably long part of the sentence with the reference itself that identifies it for within-text presentation (Fig. 1D).

### B. Presentation of document references

Document references are recognized by the extractors, and are submitted to IDOL for storage and indexing. They are presented to the user as tables, with the Celex number transformed to an internal link to the document [6].

Document content is decorated by the document presentation back-end so that references will be links to the document, and references not found in the repository will be colored red for easy identification (Fig. 1A).

Documents not found in the repository are marked up with a red font color given (Fig. 1F).

### C. Relative references

It is usual in legal texts to start with an exact reference, then replace further occurrences with a textual form (e.g. "Article 9 of Directive 2016/2284 (NEC-Directive)"). Capturing these "further" forms makes it possible to transform all further occurrences to links (Fig. 1F).

### D. Acronyms and short names

A similar situation is an exact name of e.g. an institution at first reference, then an abbreviation or acronym for further occurrences. These are best recognized when all document references are replaced by GUIDs in the text, as they may also be part of document references (e.g. "European

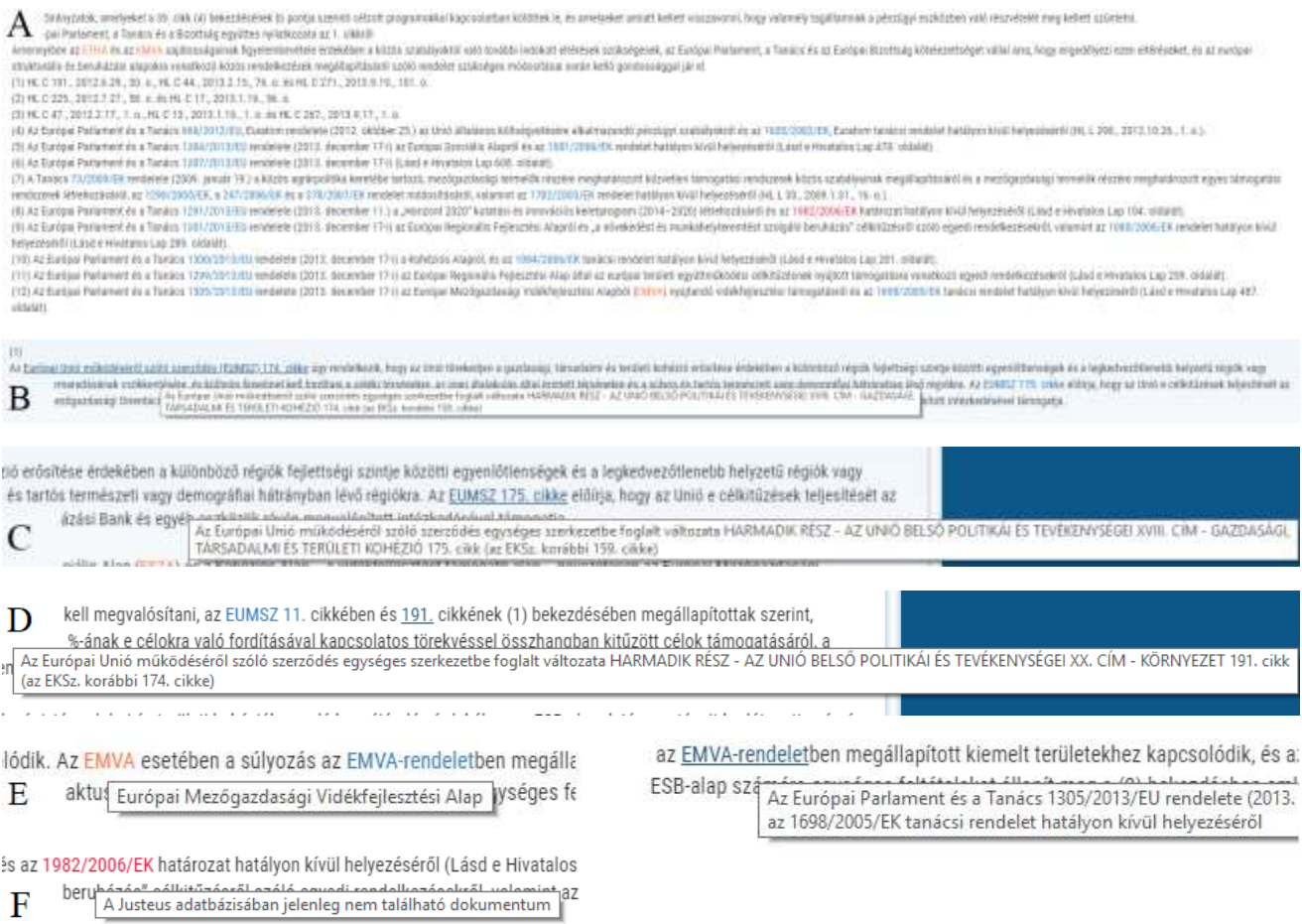

Fig. 1. Recognition and presentation of document connections and acronyms. "EMVA:" European Agricultural Fund for Rural Development; "Rendelet:" Regulation; "EUMSZ:" Treaty on the Functioning of the European Union; "cikk:" article.
A. An extensively decorated document section.
B. Full form of a EU treaty with article reference.
C. Short form of a EU treaty with article reference.
D. References for two articles of the same treaty.
E. A standalone acronym and its document reference form.
F. A document not found in the Justeus database.

Agricultural Guarantee Fund (EAGF)" referring to the institution, and "EAGF-Directive" referring to a legal document) (Fig. 1E–F).

### E. Named entitiy recognition

Hungarian court rulings, as published by the National Office for The Judiciary, at present (May 2019) provide a limited set of metadata including previous and subsequent decisions as well as cited legislation. Names of judges, applicants, defendants, their representatives and case subjects are extracted wherever available and left intact by anonymization. Judge names are extracted with the help of a normalized list, allowing one mistyped character per word. Case subjects' variability is enormous, resulting in approx. 20,000 different expressions in 150,000 documents. Normalization is therefore essential and requires expert help.

## V. THE NETWORK APPROACH TO DOCUMENT CONNECTIONS

Many documents in the Justeus repository have numerous references to other documents, the maximal actual value being around 450 at present. Most of these connection types are important to the user, and documents that have many incoming references are probably important documents [7]. If, for example, an EU directive has many references from EU and national judgements, it probably regulates a highly disputed, perhaps difficult to understand area of common interest; examples are weekly working hours or value added tax. Some other references may be less important like the obligatory mentioning of at least one European treaty in almost all EU documents, without further (article level) specification (sort of a general legal base).

Such a high number of document connections may be difficult to analyze in tabular format, especially when secondary and further order references (references in the referred documents) are also to be considered. The number of dimensions in a table is severely limited.

### A. Legal citation networks

Network graphs have been increasingly used to explore complex relationships among data. More relevant to our present subject, they have been used to study U.S. legislation complexity [8] and legal precedent evolution [9].

In the U.K., attempts have been made to identify excessively complex legislation [10].

The EU legislation corpus has also been subjected to scientific network analysis [11]. EuCaseNet [12], an application presenting network graphs, aims at EU case law only.

Our approach is at present limited to a citation network of two levels (documents referred from documents referred from the central document), including all EU legislation and case law documents as well as HU case law, presented as network graphs to the user.

Mapping graph nomenclature to our application, documents (actually document dossiers, see below) become nodes (vertices), and grouped connections become edges (links). Edges can be directed or not directed, depending on legal meaning and importance.

Real time display limits processing time and forces us to make compromises and simplifications when preparing data and rendering graphics. Data (document connections) are organized at the back-end server, while layout and rendering takes place at the client machine (browser).

### B. Document versions and dossiers

To reduce the number of nodes (documents) to layout and display, the network preparation algorithm joins related documents into a single node (dossier). Still, a lead document has to be defined to provide the label of the node and to jump to if the user clicks the node. This logic is collection dependent.

For EU treaties, the dossier collects all versions of the same article, independent of the year of publication. The lead document is the latest version.

For EU legislation, the dossier collects all versions including consolidated versions, modifications and corrigenda. The lead document is the original (earliest) version.

For EU case law, the dossier collects all documents belonging to the same case number (or the joined cases), from case announcement (application) to the final judgement and its summary, including any appeal. There are also less-defined document types like "Judicial information." The lead document is the latest.

For AB, the decision replaces the original initiative so that the dossier essentially contains a single document, except for joined cases.

For OBH, the concept of a case across courts is not defined in Hungary; a case is meant to contain all documents from application to decision within the same court. What we see is a series of court decisions following the appeals. For the time being, these documents are displayed as separate nodes.

### C. Connection pairs and importance

Connections between two nodes are displayed as single lines. However, many connection types are defined as pairs. For example, a court decision has a connection property that represents the annulation of a legal act; the legal act, in turn, has a connection property that represents the fact that it has been annulated. This redundancy facilitates search and filtering as legal acts annulated by courts can be filtered, and court decisions that annulated legal acts can also be filtered; and offers added security if one of the pair is missing. If both are present, however, it should be decided which one is displayed.

Of the two connections forming a pair, the active one is chosen for display, if present. The active connection is that affecting its target document, as opposed to a passive acknowledgement of the action. A court decision is active if it affects (modifies, suspends, annuls, confirms etc.) a directive; the directive passively registers this fact. In other cases, e.g. mentioning or citing a legal act, the connection is weaker and may not have a definite direction (e.g. "related documents").

Besides pairs, two documents may very well have multiple connections (e.g. a court decision annulated an article of a given directive, and approved another article of the same directive.) Again, a lead connection has to be defined to determine the color and arrow direction of the connection (edge; this priority goes to the more active or important connection, as determined by legal experts.

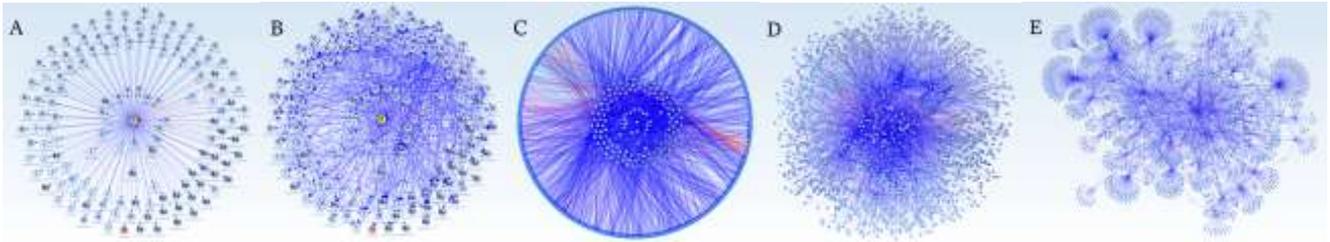

Fig. 2. Evolution of a document network graph in Justeus. A. Initial star-like formation of the document references (first order) from the central document (yellow). B. Cross-references of first order documents displayed. C. Second-order documents (documents referred to by first-order documents) displayed. D. Self-organizing diagram at an early stage of development. E. Final network display. Self-organization took approx. 1 min on a high-end PC in Chrome.

### D. Graph Layout and Rendering

The intended graph is a hierarchical layout, but calculating the screen positions of the nodes takes time. To improve user experience, graph layout calculation starts immediately when the user loads a document. The first version to present is a simple asterisk with the chosen document in the center, and documents referred from the central document arranged in one or more circles, depending on the number of nodes (Fig. 2A). Then, the user can choose to display the cross-connections between the first-level nodes (Fig. 2B). The second level of nodes is displayed again upon user request (Fig. 2C); this may lead to cumbersome graphs, resolved by a dynamic, self-organizing layout that can be viewed during its development (Fig. 2D–E).

### E. Subgraphs and User-Defined Filtering

The user can filter the nodes by their type, determined by the collection they are stored in the document repository (e.g. EU Treaties or EU Directives). Connections (edges) can also be filtered according to their types (Fig. 3).

## VI. Conclusions and Future Directions

Justeus is in beta stage and evolving. At the time of this writing, user feedbacks are being collected and search logs analyzed. It is already clear that to serve users, the HU legislation corpus should be integrated. HU legislation texts are available in simple textual form (as opposed to the legislative XML of EUR-Lex [13], NormeinRete [14] in Italy, LexDania in Denmark, CHLexML in Switzerland and eLaw in Austria). HU legislation consolidated versions are not available in the public domain for reuse.

In addition, document clustering by subject [15] and search for similar documents are desirable. At least in Hungarian, these functions may require extensive stemming, data cleaning and normalization.

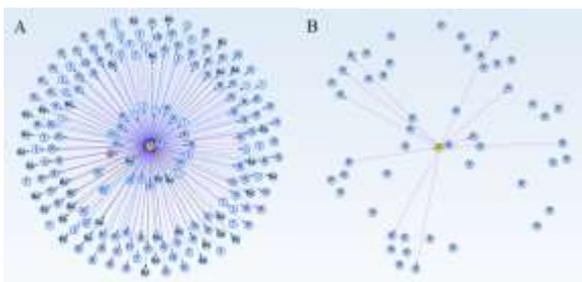

Fig. 3. Complete (A) and filtered (B) displays of the same primary document connection set. Filtering left EU regulation nodes only.